\documentclass[sigconf]{acmart}
\usepackage[utf8]{inputenc}

\usepackage{booktabs} 

\usepackage{subfig}
\usepackage{graphicx,color}
\usepackage{colortbl}
\usepackage{algorithm}
\usepackage{algorithmic}

\usepackage{bm}
\usepackage{amsmath}
\usepackage{soul}
\usepackage{rotating}
\usepackage{amssymb}
\usepackage{amsfonts}
\usepackage{multirow}
\usepackage{balance}
\usepackage{verbatim}
\usepackage{paralist}
\usepackage{enumitem}
\setlist{leftmargin=4mm}

\usepackage[T1]{fontenc}

\usepackage{tabularx}
\newcolumntype{Y}{>{\centering\arraybackslash}X}

\newcommand{\ignore}[1]{}

\newcommand{\todo}[1]{\textit{\textcolor{orange}{(#1)}}}
\newcommand{\mar}[1]{\textcolor{magenta}{#1}}

\setcopyright{none}

\acmDOI{10.1145/nnnnnnn.nnnnnnn}

\acmISBN{978-x-xxxx-xxxx-x/YY/MM}

\acmConference[]{}{}{}
\acmYear{2019}
\copyrightyear{2019}

\acmPrice{15.00}

\usepackage{color}

\begin{document}
\title{A characterisation of S-box fitness landscapes in cryptography}

\author{Domagoj Jakobovic}
\affiliation{%
  \department{}
  \institution{University of Zagreb}
  \streetaddress{}
  \city{} 
  \state{Croatia} 
  \postcode{}
}
\email{}

\author{Stjepan Picek}
\affiliation{%
  \department{}
  \institution{Delft University of Technology}
  \streetaddress{}
  \city{} 
  \state{The Netherlands} 
  \postcode{ }
}
\email{}

\author{Marcella S. R. Martins}
\affiliation{%
  \department{}
  \institution{Federal University of Technology}
  \streetaddress{}
  \city{} 
  \state{Brazil} 
  \postcode{}
}
\email{}

\author{Markus Wagner}
\affiliation{%
  \institution{University of Adelaide}
  \streetaddress{}
  \city{}
  \state{Australia} 
  \postcode{}
}
\email{}

\renewcommand{\shortauthors}{Jakobovic et al.}

\begin{abstract} 

Substitution Boxes (S-boxes) are nonlinear objects often used in the design of cryptographic algorithms. The design of high quality S-boxes is an interesting problem that attracts a lot of attention. Many attempts have been made in recent years to use heuristics to design S-boxes, but the results were often far from the previously known best obtained ones. Unfortunately, most of the effort went into exploring different algorithms and fitness functions while little attention has been given to the understanding why this problem is so difficult for heuristics. In this paper, we conduct a fitness landscape analysis to better understand why this problem can be difficult. Among other, we find that almost each initial starting point has its own local optimum, even though the networks are highly interconnected.

\end{abstract}

%
%



\maketitle


\section{Introduction}
\label{sec:introduction}

One of the key goals in cryptography is the design of new ciphers. A common choice to encrypt traffic is to use block ciphers, which are deterministic algorithms operating on fixed length groups of bits and with an unvarying transformation specified by a symmetric key.
To design a block cipher, the designer has several directions he can follow but one widely accepted choice is to use 
Substitution-Permutation Network (SPN) cipher. Such ciphers usually consist of an XOR operation with the key/subkeys, a linear layer, and a substitution layer~\cite{knudsen}.  

To build the substitution layer, a common
option is to use one or more Substitution Boxes
(S-boxes, vectorial Boolean functions, $(n, m)$-functions). 
Such S-boxes need to fulfill certain cryptographic properties in order to provide resilience against cryptanalysis.
More precisely, an S-box in SPN should be bijective (so, $n=m$), with high nonlinearity, and low differential uniformity (just to name a few of the most important properties). 
Additionally, in order for ciphers to be used in various environments, they use different sizes of input and consequently different S-boxes. Today, common choices for bijective S-box sizes are $4\times 4$ (PRESENT~\cite{present07}),
$5\times 5$ (Keccak~\cite{keccak}), and $8\times 8$ (AES~\cite{Daemen}).
To obtain such S-boxes, one has several options: 1) algebraic constructions~\cite{CramaSbox}, 2) random search, and 3) heuristics.

When considering heuristics for the construction of S-boxes, the current results are not so promising. Except for the smallest sizes (e.g., up to $4\times 4$), such techniques are not able to find highly fit S-boxes. Better results are obtained only in cases where one either does not start from a random solution but highly fit solutions, or when some restrictions are considered in order to reduce the complexity of the search. Consequently, in most of the scenarios, there does not seem to be a clear advantage in using heuristics over algebraic constructions. An exception is when one considers some properties that result in suboptimal values when using algebraic constructions, but it is often not clear whether such properties are really relevant. Still, we consider this to be an interesting problem and one where, if further improvements are made, heuristics can play a significant role. 

\begin{table}[t]
\centering
\small\vspace{3mm}
 \begin{tabular}{cccccc}\toprule
 	$n\times n$ & $3\times 3$ & $4\times 4$ & $5\times 5$ & $6\times 6$ & $7\times 7$\\\midrule
 	Size & $8! \approx 2^{15}$ & $16! \approx 2^{44}$  & 
 	$32! \approx 2^{117}$  & $64! \approx 2^{296}$ & $128! \approx 2^{716}$  \\\midrule
 	$N_F$ & 2 & 4 & 12 & 24 & 56\\\bottomrule
 \end{tabular}
\caption{S-boxes, search space sizes and max nonlinearities.}
\label{tab:sboxes}\vspace{-4mm}
\end{table}

Intuitively speaking, there are at least two reasons for the difficulty of finding bijective, highly nonlinear S-boxes: 1) the search space size, and 2) the lack of information to guide the search.
In total, there are $2^{m2^n}$ S-boxes, out of which $2^n!$ are bijective. Already for a relatively small size, this represents a huge search space.
When considering the lack of the information to guide the search, we need to first decide how to represent the S-box. A trivial choice would be to represent it as a number of Boolean functions (since an S-box is a vectorial Boolean function) but the results in literature suggest this approach is not adequate for obtaining bijective S-boxes for sizes larger than $4\times 4$~\cite{Picek:2015:CGP:2739482.2764698}.
To alleviate this problem, a common choice is to use permutation encoding where bijectivity is inherently preserved. When considering the nonlinearity property, for bijective S-boxes it assumes values in steps of 2 (i.e., nonlinearity of 0, 2, 4, etc.). As a consequence, we can imagine that there is simply not enough information in the search space to converge to better values.
Still, the question remains whether these two issues are the only problems one would encounter or is there something more.
In Table~\ref{tab:sboxes}, we give the search space sizes for several S-boxes dimensions, as well as the maximal obtainable nonlinearity. We give details on how those values are obtained in Section~\ref{sec:sboxes}.

Up to now, the research for new S-boxes has  been mainly directed into developing new, improved fitness functions or selecting more appropriate encodings. Little, if any, effort has been done in trying to understand why this problem is difficult and how to improve the performance of commonly used heuristic techniques.
In this paper, we conduct fitness landscape analysis (FLA), using a simplified landscape representation called Local Optima Networks (LONs)~\cite{ochoa2008study}. It uses a graph whose nodes are associated with local optima (or basins of attraction) and where edges indicate connectivities between the local optima. The search space can be exploited using existing graph theories, making FLA a very useful tool for combinatorial problems.

\section{Related Work}
\label{sec:related}

When considering the results on S-boxes obtained with heuristics, there are certain observations one can make. 
First, we can see that most of the works consider evolutionary algorithms and the permutation encoding. Next, we can see that the results obtained with heuristics are usually subpar when compared to the results obtained with algebraic constructions.
Finally, to the best of our knowledge, all such papers consider how to obtain better solutions while there is no attempt to try to understand why is this problem so difficult for heuristics.

Clark et al. used the principles from the
evolutionary design of Boolean functions to evolve S-boxes with the desired cryptographic properties for sizes up to $8\times 8$~\cite{Clark_thedesign}.
Burnett et al. used a heuristic method to generate  S-boxes to be used in MARS cipher~\cite{Burnett2}. With this approach, the authors were able to generate a number of S-boxes of appropriate sizes that satisfy all the requirements placed on a MARS S-box. 
Picek et al. used Cartesian Genetic Programming and Genetic Programming to evolve S-boxes and discussed how to obtain permutation based encoding with those algorithms~\cite{picekgeccop32015}. 
Picek et al. presented an improved fitness function with which EC is able to find higher nonlinearity values for a number of S-box sizes~\cite{picek2016new}. 
Picek et al. also discussed how to use genetic programming to evolve cellular automata rules that in turn can be used to generate S-boxes with good cryptographic properties~\cite{Picek2017a,Picek2017b}. Interestingly, the results obtained in these two papers, where GP is used to evolve CA rules, outperform any other solutions obtained with heuristics for sizes $5\times 5$ up to $7\times 7$.


\section{Substitution Boxes}
\label{sec:sboxes}

In general, an S-box takes $n$ input bits and transforms them into $m$ output bits, where $n$ is not necessarily equal to $m$. In the following, we cover the mathematical preliminaries on S-boxes, and we define the fitness functions used in this study.

\subsection{Mathematical Preliminaries}

Let $n, m$ be positive integers, i.e., $n, m \in \mathbb{N}^+$. We denote by
$\mathbb{F}_{2}^{n}$ the $n$-dimensional vector space over the finite field
$\mathbb{F}_{2}$. Further, for any set $S$, we denote $S \backslash \{0\}$ by
$S^{*}$.  The usual inner product of $a, b \in \mathbb {F}_2^n$ equals $a\cdot b = \bigoplus_{i=1}^{n} a_{i} b_{i}$.

An $(n,
m)$\nobreakdash-\hspace{0pt}function is any mapping \textit{F} from
$\mathbb{F}_{2}^{n}$ to $\mathbb{F}_{2}^{m}$.  An $(n, m)$-function \textit{F}
can be defined as a vector $F = (f_1,\cdots,f_m)$, where the Boolean functions
$f_i: \mathbb{F}_2^n \rightarrow \mathbb{F}_2$ for $i \in \{1, \cdots, m\}$ are
called the coordinate functions of $F$. Given $v \in (\mathbb{F}_2^m)^*$, the
component function $v\cdot F:\mathbb{F}_2^{n} \rightarrow \mathbb{F}_2$ is the Boolean
function defined for all $x \in \mathbb{F}_2^n$ as the inner product between $v$ and
$F(x)$. In other words, the component functions of $F$ represent the non-trivial
linear combinations of its coordinate functions.

The Walsh-Hadamard transform of an $(n, m)$-function $F$ is defined as
(see e.g.,~\cite{CramaSbox}):
\begin{equation}
\label{eq:walshHadamard}
W_{v\cdot F} (\omega) = \sum_{x \in \mathbb{F}_{2}^{n}} (-1)^{v\cdot F(x) \oplus
  \omega \cdot x}, \ v \in (\mathbb{F}_{2}^{m})^*, \ \omega \in \mathbb{F}_2^n \enspace .
\end{equation}

In particular, the quantity $W_{v\cdot F} (\omega)$ measures the correlation  between the component function $v\cdot F$ and the linear
function $\omega\cdot x$. 

In order to resist linear cryptanalysis, an S-box should ideally have \emph{high nonlinearity}.  
Also, when used in cryptographic algorithms following the Substitution Permutation Network (SPN) design principle, S-boxes also need to be bijective.

A $(n, m)$-function $F$ is balanced if it takes every value of
$\mathbb{F}_{2}^{m}$ the same number $2^{n - m}$ of times. Balanced
$(n,n)$-functions correspond to bijective S-boxes (i.e, permutations on $[0, 2^n-1]$).

The nonlinearity $N_F$ of an $(n, m)$-function  \textit{F} equals:
\begin{equation}
\label{eq:nonlinearity_sbox}
N_F = 2^{n - 1} - \frac{1}{2}\max_{\substack{\text{$\omega \in \mathbb{F}_{2}^{n}$}\\\text{$v \in
      \mathbb{F}_{2}^{m*}$}}} |W_{v \cdot F} (\omega)|.
\end{equation}

The nonlinearity of any $(n, n)$ function $F$ is bounded above by the so-called Sidelnikov-Chabaud-Vaudenay bound~\cite{Chabaud1995}:
\begin{equation}
\label{eq:max_nl}
N_F \leq 2^{n-1} - 2^{\frac{n-1}{2}}.
\end{equation}
Bound (\ref{eq:max_nl}) is an equality if and only if $F$ is an Almost Bent  (AB) function, by definition of AB functions (note, AB functions exist in only odd number of variables)~\cite{CramaSbox}.
When $n$ is even, the best possible nonlinearity value is believed to be equal to $2^{n-1} - 2^{\frac{n}{2}}$ (for $n \leq 6$ it is experimentally confirmed).

As already said, the nonlinearity values change in steps of two. This is a consequence of the fact that the values of the Walsh-Hadamard spectrum are even and divisible by $2^{t+1+\left\lfloor \frac{n-t-1}{d}\right\rfloor}$ where $n$ is the dimension of coordinate Boolean function, $t$ is its correlation immunity ($t \leq n - 2$ and is non-negative integer), and $d$ is the algebraic degree~\cite{10.1007/3-540-44495-5_3}. Note that by the Siegenthaler bound $d \leq n - t - 1$~\cite{CramaBoolean}.

\subsection{Fitness Functions for Optimisation of S-boxes}

Among many choices, in this paper we consider two fitness functions. The first one simply uses the nonlinearity value, and the goal is to maximise it:
\begin{equation}
\label{eq:fit_NL}
    NL = N_F.
\end{equation}

In the second fitness function, motivated by an observation made in  \cite{picek2016new}, we additionally use the information about how many \textit{component functions} of the S-box have the smallest nonlinearity (since the smallest nonlinearity is the final nonlinearity).
Intuitively, by minimising the number of such smallest values we increase our chances to improve the nonlinearity (if we are able to remove all the components with the smallest value). Note, this fitness function does not consider the distribution of nonlinearity over all component functions.
Although in related work one can find several ways how to increase the granularity of fitness functions based on the nonlinearity value, to the best of our knowledge, the technique with component functions was not used before.

Note, since this is just an indicator of a behavior and not a reliable measure (because the number of the components with the smallest value does not mean anything without knowing the smallest value), we encode it in the range $[0,1]$ as $Indicator = 1/(\# \ smallest \ nonlinearity \ value)$.
Finally, the second fitness function, denoted with $NL_f$, aims to maximise the following expression:
\begin{equation}
\label{eq:fit_NLf}
    NL_f = N_F + Indicator.
\end{equation}


\section{Fitness Landscape Analysis}

Fitness landscapes illustrate the association between search and fitness space~\cite{kauffman1987towards,kauffman1993structure},  
defined as the triple $(S,F,N)$ where $S$ is the search space, $F$ the objective function, and $N$ the neighbourhood operator~\cite{watson2010introduction}. 
A heuristic can be seen as a strategy of search in this structure for near-optimal solutions~\cite{tavares2008multidimensional}.

Fitness landscape analysis (FLA) has been applied to investigate the dynamics of local search techniques, evolutionary algorithms, single-solution heuristics and other metaheuristics~\cite{watson2005linking} for optimisation and design problems~\cite{verel2011localNK,MartinsYSDLA18moEDA, Yafrani2018FLA}, using models to predict the behaviour of these techniques~\cite{watson2005linking}. 
The behaviour is generally illustrated by the cost required to locate a solution with a given quality threshold given a problem instance. These models can also identify which features
of the fitness landscape are responsible for the problem difficulty during the search process~\cite{watson2010introduction}. By identifying these features, algorithm performance can be improved.  

Multiple fitness landscape models have been applied for combinatorial problems. The Local Optima Networks (LONs), proposed by \citet{ochoa2008study}, is a simplified model designed to understand the structural organisation of the local optima in combinatorial landscapes.

In LON models the fitness landscape is represented as a graph where the
nodes are the local optima that can be connected. 
A local search heuristic $\mathcal{H}$ defines a mapping from the solution space $S$ to the set of locally optimal solutions $S^*$. 
A solution $i$ in the solution space $S$ is a local maximum given a neighbourhood operator $N$ if $F(i)\geq F(s), \forall s \in N(i)$. 

Each local optima $i$ has an associated basin of attraction set defined by $B_{i}=\{s\in S | \mathcal{H}(s)=i\}$. This set is composed of all the solutions that, after applying a local search procedure starting from each of them, the procedure returns $i$. The size of the basin of attraction of $i$ is the cardinality of $B_{i}$. Given a neighbourhood operator, there is a connection between two attraction basins if at least one solution in one basin has a neighbour solution in the other basin. 

\begin{figure}[t]
\vspace{-3mm}
\centering
\subfloat{
  \centering
  \includegraphics[scale=.2]{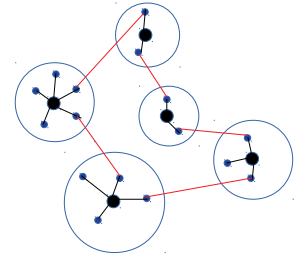}
}
\vspace{-2mm}
\caption{
A simplified illustration of the attraction basins and the connectivity in local optima networks.\label{fig:basins-lon}
}
\vspace{-1mm}
\end{figure}

Figure \ref{fig:basins-lon} illustrates a simplified LON, for visualisation purposes, including the basin of  attraction (blue circles), their local optima (black big dots), the solutions that converge to the local optima when applying the local search (blue small dots), and the edges between the local optima (red lines) that exist due to neighbour-relationships. 
It is important to note that sophisticated heuristics are likely 
expected to result in more interconnections between the basin of attraction than what we have shown in the illustration.

Local optima network properties for permutation-based problems have been studied in \cite{daolio2010local} and \cite{daolio2013local}. Additionally, some works investigated the correlation between LON features and the performance of 
search heuristics~\cite{verel2008connectivity,ochoa2010first,chicano2012local,daolio2012local,ochoa2014local}.

In~\cite{verel2011localESCAPE}, the authors extended the LON formalism to neutral fitness landscapes. Neutral networks are connected networks of genotypes of equal (or quasi equal) fitness, with sporadic jumps between them. The study is based on two neutral variants of the NK landscape model, in which the amount of neutrality can be tuned. The results from the 
addressed network features confirmed that the study of neutrality may enhance heuristic search.

More recently,~\cite{ochoa2017mapping} applied the LON model to characterise and visualise the global structure of travelling salesperson (TSP) fitness landscapes of different classes including random and structured real-world instances of realistic size. The results revealed that the landscapes of the tested instances have multiple valleys or funnels, which are related to search difficulty on the studied landscapes. The authors also provided ways of analysing and visualising the neutral landscapes featured by the structured TSP instances.

The work presented by \cite{hernando2017local} considered a LON variant called Compressed Local Optima Network, which was used to investigate different landscapes for the Permutation Flowshop Scheduling Problem (PFSP). The authors analysed the network features in order to find differences between the landscape structures, giving insights into which features impact the performance of an Iterated Local Search heuristic.

In \cite{Yafrani2018FLA} the authors investigated two hill climbing local search procedures and the corresponding LONs were explored to understand the difficulty of Travelling Thief Problem (TTP) instances. Some problem features were extracted and graphs with statistical analysis were presented. Among other, they found that certain operators can result in LONs with disconnected components, and that at times exploitable correlations of node degree, basin size, and fitness exist.

In this paper we investigate a greedy, deterministic local search exploring different types of neighbourhood and fitness functions. A LON model is designed and adapted for the optimisation of S-boxes. This way, we can understand the impact of some problem features, using measures that assess the connectivity and attraction basins, as well as the statistical tests to explore the scale-freeness of the obtained LONs.


\section{Local Search Heuristics}

In this section, we present the local search framework for the optimisation of S-boxes.
The employed local search is a greedy deterministic hill climber, designed with the primary goal of investigating the structure of the search space. 
The pseudocode is described in Algorithm \ref{algo:ls}; the algorithm can be used with an arbitrary representation and an arbitrary neighbourhood relationship, where $\mathcal{N}(.)$ represents the neighbourhood of the given solution.
Note that the algorithm is deterministic; if there are multiple solutions within the neighbourhood with the same fitness value, the algorithm will select the first one that it encounters.
The ordering of the solutions in the neighbourhood depends on the actual neighbourhood relation.

\begin{algorithm}[ht]
\small
\caption{A greedy local search heuristic\label{algo:ls}}
\begin{algorithmic}[1]
  \STATE $s \gets $ initial solution
  \WHILE{there is an improvement}
    \STATE $s^* = s$
    \FOR{each $s^{**}  \text{ in } \mathcal{N}(s)$}
        \IF{$F(s^{**}) > F(s^*)$}
          \STATE $s^* \gets s^{**}$
        \ENDIF
    \ENDFOR
    \STATE $s = s^*$
  \ENDWHILE
\end{algorithmic}
\end{algorithm}

For the S-box representation, we use the permutation encoding which preserves the bijectivity property; in our experiments, each individual is a permutation of size $2^n$ for a $n \times n$ S-box.
In the context of this study, we consider two local neighbourhood variants with Algorithm \ref{algo:ls}:

\begin{enumerate}
\item The first (denoted ``swap'') uses the swap (or also called ``toggle'') operation to generate the neighbourhood; the swap operation takes two different positions in the permutation vector and exchanges them.
\item The second variant (denoted ``invert'') uses the \emph{inversion} operator; this operator inverts the ordering of permutation elements between two different boundary points. 
\end{enumerate}

The size of the explored neighbourhood is the same for both operators; the two positions in the permutation vector ($i$ and $j$) assume the values in range $i\in \left\{ 1, ..,  2^n-1 \right\}$ and $j\in \left\{ i+1, ..,  2^n \right\}$.
Total number of combinations is
\begin{equation}
    \sum_{i=1}^{2^n-1} \sum_{j=i+1}^{2^n} 1 = \sum_{i=1}^{2^n-1}2^n - \sum_{i=1}^{2^n-1}i = \frac{2^n\left ( 2^n-1 \right )}{2}
\end{equation}
for a permutation of size $2^n$. 
For example, this means that in the case of investigating 7x7 S-boxes, each search point has 8127 neighbours that need to be investigated in lines 4--8 of Algorithm~\ref{algo:ls}.

For every initial solution, the local search procedure in Algorithm~\ref{algo:ls} is performed until no more improvement is possible; this results in a single local optimum.
Along with the optimum, a chain of solutions leading from the initial solution to the local optimum is added as the members of the local optimum's basin of attraction.

When exploring the neighbourhood, the implementation keeps track of all the intermediate solutions that lead to a local optimum.
During the algorithm execution, each new solution being considered is compared to already visited solutions so that every local optimum remains unique:
\begin{compactitem}
    \item if the currently considered solution already appears in an existing basin, local search is terminated, and all the solutions leading to the current one are added to the existing basin;
    \item only if every solution in the search sequence from the initial solution to the local optimum has not been encountered before, a new local optimum with its basin is created.
\end{compactitem}

After all the initial solutions have been explored by the local search, in the second phase we consider every local optimum basin and find connections between basins.
If any solution from one basin is a neighbour of any solution in the second basin, the connection is formed and the same procedure is repeated for every pair of basins.

\section{Experiments and Results} 

In this section, we analyse the local optima networks obtained using the local search heuristic for the S-box optimisation problem to achieve some insights about the structure of the search space.
Furthermore, we study the basins of attraction and their relationship with some LON properties looking for additional information about the search difficulty.

In our experiments, we explore the following parameters of the search space:
\begin{itemize}
    \item{ S-box size ($3 \times 3$ and larger);}
    \item fitness function: $NL$ (Equation~\ref{eq:fit_NL}) or $NL_f$ (Equation~\ref{eq:fit_NLf});
    \item{ neighbourhood type (swap, invert);}
    \item{ number of samples (unique initial solutions).}
\end{itemize}

For problem size $3 \times 3$, it is possible to perform an exhaustive search, since the total number of solutions is $8!=40,320$.
In larger sizes, we employ a fixed sample size, which is the number of unique initial solutions; for each solution, we run Algorithm~\ref{algo:ls} until no further improvements are possible.
As the search space in the problem at hand is enormous, generating purely random initial solutions leads to a network of local optima that are almost never connected, since they are too far away in the search space.
To get an insight into the structure of the explored neighbourhood relation, we generate initial solutions as lexicographically ordered permutations, starting with a random permutation in each experiment.

\subsection{Topological properties of local optima networks}

Tables~\ref{tab:metricsLON3x3} and \ref{tab:metricsLON} report on properties that are often used for LON analyses~\cite{ochoa2008study}. 
We extract some graph metrics as follows.
$n_v$ and $n_e$ represent the number of vertices (or nodes) and the number of edges of the generated LON respectively.  
$z$ is the average degree.
$C$ is the average clustering coefficient.
$C_r$ is the average clustering coefficient of corresponding random graphs (i.e. random graphs with the same number of vertices and mean degree).
$l$ is the average shortest path length between any two local optima.
$\pi$ is the connectivity, which represents if the LON is a connected graph.
Finally, $S$ is the number of non-connected components (sub-graphs). For \textit{invert}, we have typically not reached $100,000$ samples due to the higher computational cost of performing each inversion on solutions of length $n!$. These cost result in a multiplicative factor of $\Theta(n!)$ to the time required for the complete neighbourhood searches.

\begin{table*}[h!]
\small
\centering
\begin{tabularx}{\ignore{\textwidth}16cm}{ *{10}{Y}}
\toprule
\ignore{\textbf{Size}    &} \textbf{Function} &\textbf{Operator} & \textbf{${n_v}$} & \textbf{${n_e}$} & \textbf{${z}$} & \textbf{${C_r}$} & \textbf{${C}$} & \textbf{${l}$}    & \textbf{$\pi$} & \textbf{${S}$} \\ \hline

\ignore{3x3NLswap	&}\multirow{2}{*}{$	NL	$}&$	swap	$&$	10,752	$&$	169,344	$&$	31.5000	$&$	0.0029	$&$	0.0748	$&$	3.6373	$&$	1.00	$&$	1.00	$\\
\ignore{3x3NLinvert	&} &$	invert	$&$	10,752	$&$	593,376	$&$	110.375\ignore{110.3750}	$&$	0.0103	$&$	0.0947	$&$	2.5466	$&$	1.00	$&$	1.00	$\\
\ignore{3x3NLfswap	&}\multirow{2}{*}{$	NL_f	$}&$	swap	$&$	10,752	$&$	203,616	$&$	37.8750	$&$	0.0035	$&$	0.1044	$&$	3.5359	$&$	1.00	$&$	1.00	$\\
\ignore{3x3Nlfinvert	&}&$	invert	$&$	10,752	$&$	657,888	$&$	122.375\ignore{122.3750}	$&$	0.0114	$&$	0.1006	$&$	2.4918	$&$	1.00	$&$	1.00	$\\
\bottomrule
\end{tabularx}
\caption{General LON and basins' statistics for S-box  $3 \times 3$.
}
\label{tab:metricsLON3x3}\vspace{-2mm}
\end{table*}


\begin{table*}[h!]
\small
\centering
\begin{tabularx}{16cm}{*{11}{Y}}
\toprule
\textbf{Size}    & \ignore{\textbf{Function} &}\textbf{Operator} & \textbf{Samples} & \textbf{${n_v}$} & \textbf{${n_e}$} & \textbf{${z}$} & \textbf{${C_r}$} & \textbf{${C}$} & \textbf{${l}$}    & \textbf{$\pi$} & \textbf{${S}$} \\ \hline
\multirow{4}{*}{4x4}\ignore{4x4NLfswap20000	    &$	Nlf	$}&$	swap	$&$20,000$&$	16846	$&$	172445	$&$	20.4731	$&$	0.0013	$&$	0.0026	$&$	4.8479	$&$	1.00	$&$	1.00	$\\
\ignore{\multirow{3}{*}{4x4}4x4NLfswap100000	&$	Nlf	$}&$	swap	$& $100,000$ & $	74,641	$&$	908,454	$&$	24.3420	$&$	0.0003	$&$	0.0026	$&$	5.3995	$&$	1.00	$&$	1.00	$\\
\ignore{4x4NLfswap500000	&$	Nlf	$}&$	swap	$& $500,000$ & $	351,313	$&$	4,943,785	$&$	28.1446	$&$	0.0001	$&$	0.0035	$&$	-^\dagger	$&$	1.00	$&$	1.00	$\\
\ignore{4x4NLfinvert100000	&$	Nlf	$}&$	invert	$& $100,000$ & $	81,388	$&$	7,135,032	$&$	175.334\ignore{175.3338}	$&$	0.0022	$&$	0.3530	$&$	2.9936	$&$	1.00	$&$	1.00	$\\
\midrule
\multirow{3}{*}{5x5}\ignore{5x5NLfswap10000	    &$	Nlf	$}&$	swap	$& $10,000$ & $	7,370    $&$	65,383	$&$	17.7430	$&$	0.0023	$&$	0.0108	$&$	4.4546	$&$	1.00	$&$	1.00	$\\
\ignore{5x5NLfswap100000	&$	Nlf	$}&$	swap	$& $100,000$ & $	85,087	$&$	1,376,947	$&$	32.3656	$&$	0.0004	$&$	0.0262	$&$	4.1791	$&$	1.00	$&$	1.00	$\\
\ignore{5x5NLfinvert10000	&$	Nlf	$}&$	invert	$& $10,000$ & $	9,112	$&$	2,181,838	$&$	478.893\ignore{478.8933}	$&$	0.0526	$&$	0.6978	$&$	1.9653	$&$	1.00	$&$	1.00	$\\\hline
\multirow{3}{*}{6x6}\ignore{6x6NLfswap10000	    &$	NL_f	$}&$	swap	$& $10,000$ & $	9,676	$&$	97,447	$&$	20.1420	$&$	0.0021	$&$	0.0088	$&$	5.5936	$&$	1.00	$&$	1.00	$\\
\ignore{6x6NLfswap100000	&$	Nlf	$}&$	swap	$& $100,000$ & $	99,583	$&$	1,420,307	$&$	28.5251	$&$	0.0003	$&$	0.0010	$&$ 5.6097	$&$	1.00	$&$	1.00	$\\
\ignore{6x6NLfinvert10000	&$	Nlf	$}&$	invert	$& $10,000$ & $	9,695	$&$	1,821,963	$&$	375.856\ignore{375.8562}	$&$	0.0388	$&$	0.8029	$&$	1.9693	$&$	1.00	$&$	1.00	$\\
\midrule
\multirow{2}{*}{7x7}\ignore{7x7NLfswap10000     &$  NL_f $}&$ swap    $& $10,000$ & $ 9,998    $&$ 103,048 $&$ 20.6137    $&$ 0.0020 $&$ 0.0001   $&$ 5.0521   $&$ 1.00       $&$ 1.00  $\\
\ignore{7x7NLfinvert10000	&$	Nlf	$}&$	invert	$& $10,000$ & $	9,653	$&$	673,460	$&$	139.534\ignore{139.5338}	$&$	0.0145	$&$	0.6575	$&$	1.9901	$&$	1.00	$&$	1.00	$\\
\bottomrule
\end{tabularx}
\caption{General LON and basins' statistics. The fitness function $NL_f$ is always used. 
$^\dagger$ The average shortest path length between any two local optima could not always be computed due to the size of the networks.
}
\label{tab:metricsLON}\vspace{-2mm}
\end{table*}

In the first table,  when we perform an exhaustive search, we compare the two fitness functions $NL$ and $NL_f$ for the S-box problem of size 3x3. We find that the number of vertices ($n_v$) is the same, which indicates that the local optima and the distinct starting points are the same. 
However, the actual LONs are quite different, with the number of edges ($n_e$) ranging between about 170,000 and 660,000. 
In particular, the number of edges ($n_e$) and average degree ($z$) are lower for \textit{swap} neighbourhood type than for \textit{invert}. Besides, the $NL_f$ fitness function results in a greater number of edges and, consequently, greater average degree than $NL$. 
Interestingly, the LON consists of only one component, meaning that, in combination with the observed high mean degree and small minimum distances between nodes, a Tabu Search~\cite{glover1989tabu,glover1990tabu} with restarts or a Memetic Algorithm~\cite{Moscato1999MAS} with built-in local searches, or even an approach with explicit niching might be able to perform well and explore the entire network.

In the second table we find that the number of vertices increases with the size of S-box and when the number of samples (i.e., starting point for the local search) increases. In particular, the number of vertices in the LON increases almost linearly with the number of samples, indicating that there is a very large number of local optima in the landscape. When comparing \textit{swap} with \textit{invert}, one notices that the latter's LONs have significantly more edges, which bring down the average distance between two nodes to about 2. Possibly, this is because inversions are more disruptive than swaps. 

The clustering coefficient ($C$) of a node $i$ measures how close its neighbours are to being a clique  and it characterizes the extent to which nodes adjacent to node $i$ are connected to each other, determining, together with $l$, whether a graph is a small-world network. 
For both Tables~\ref{tab:metricsLON3x3} and~\ref{tab:metricsLON} we can observe that the LONs show a significantly higher degree of local clustering compared with their corresponding random graphs ($C_r$), except for 7x7 using \textit{swap} with 10000 samples. 
This means that the local optima are connected in two ways: dense local clusters and sparse interconnections, which can be difficult to find and exploit, specially with \textit{invert} neighbourhood.  Besides this, all LONs have a small minimal path length, i.e. any pair of local optima can be connected by traversing only few other local optima. 

Furthermore, for all cases, each local optimum is connected ($\pi=1$), meaning that there is a path between every pair of nodes. Beside, the LON is itself connected, consisting of the whole graph ($S=1$).

\subsection{Degree Distributions}


\begin{figure*}[h!]
\centering
\subfloat[]{
  \centering
  \includegraphics[scale=.28]{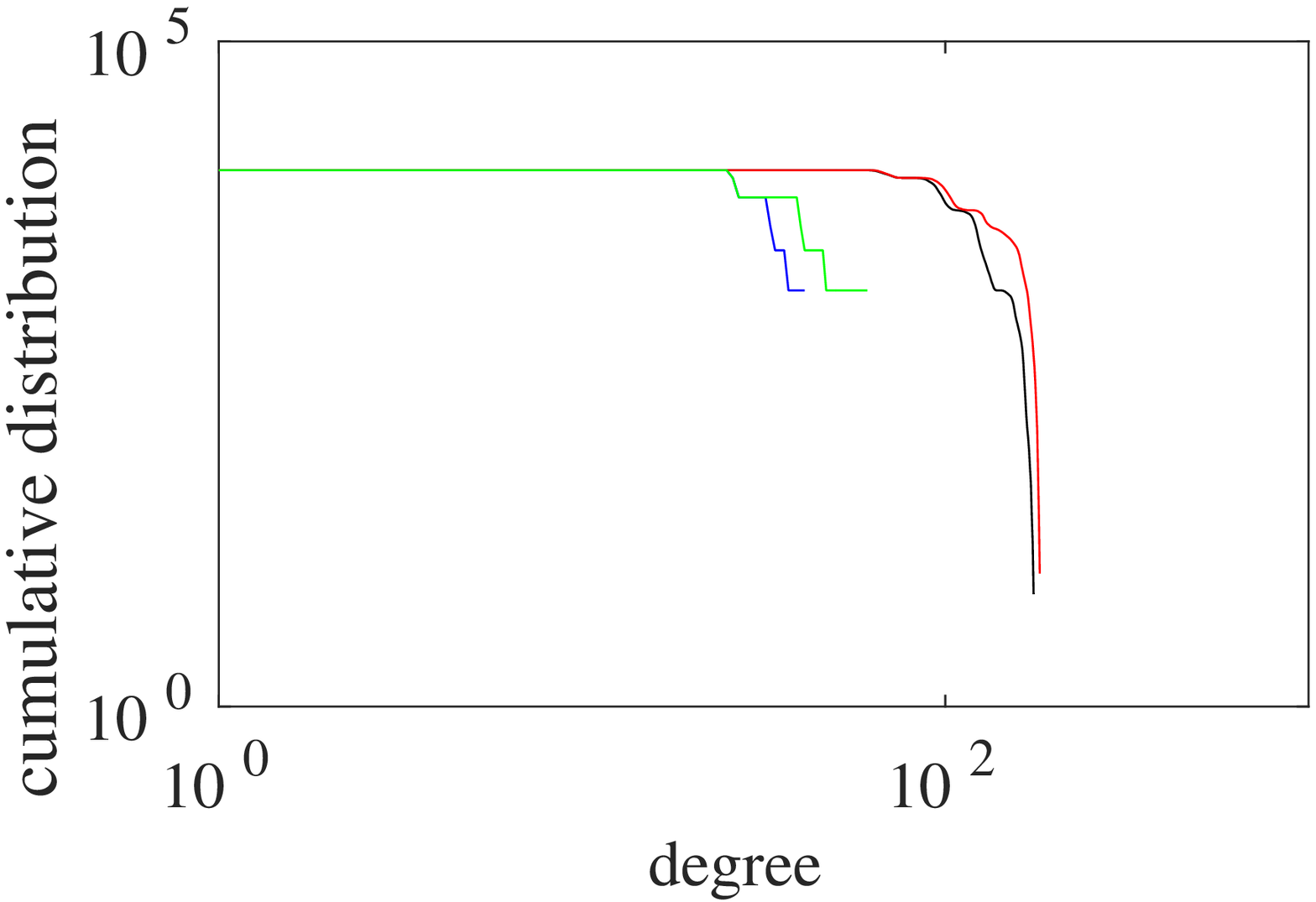}
  \label{fig:degree-dist3x3}
}
\quad
\subfloat[]{
  \centering
  \includegraphics[scale=.28]{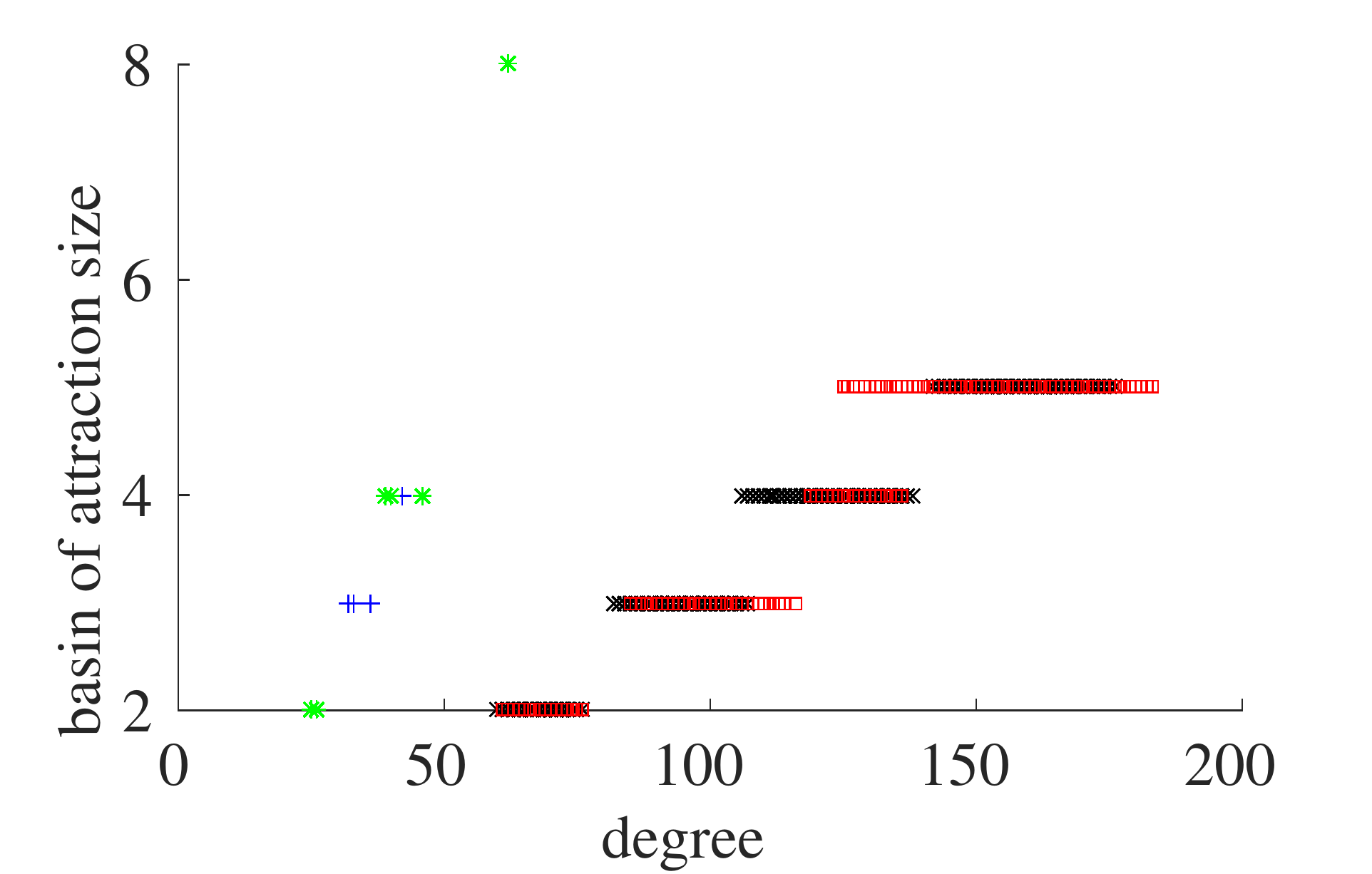}
  \label{fig:degree-basin3x3}
}
\quad
\subfloat[]{
  \centering
  \includegraphics[scale=.28]{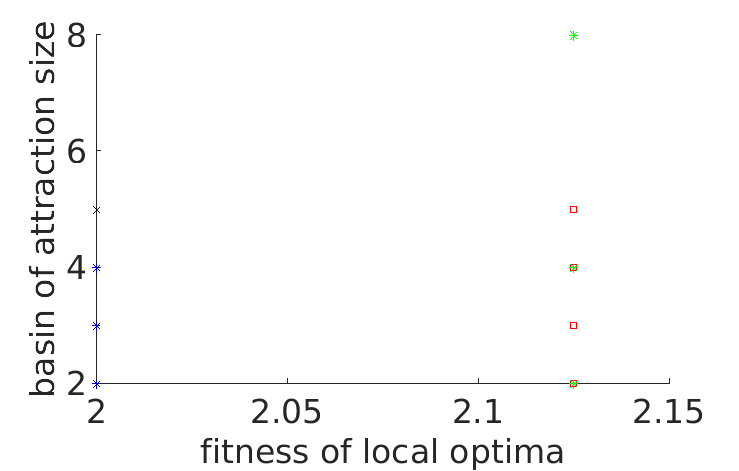}
\label{fig:basin-lofitness3x3}
}
\caption{
Statistical measures on exhaustive 3x3 $NL$ invert (black), $NL$ swap (blue), $NL_f$ invert (red)  and $NL_f$ swap (green): a) Cumulative degree distribution, b) Correlation between the degree of local optima and their corresponding basin sizes and c) Correlation between the fitness of local optima and their corresponding basin sizes.}
\label{fig:stats-measures3x3}
\end{figure*}



\begin{figure*}[h!]
\centering
\subfloat[]{
  \centering
  \includegraphics[scale=.28]{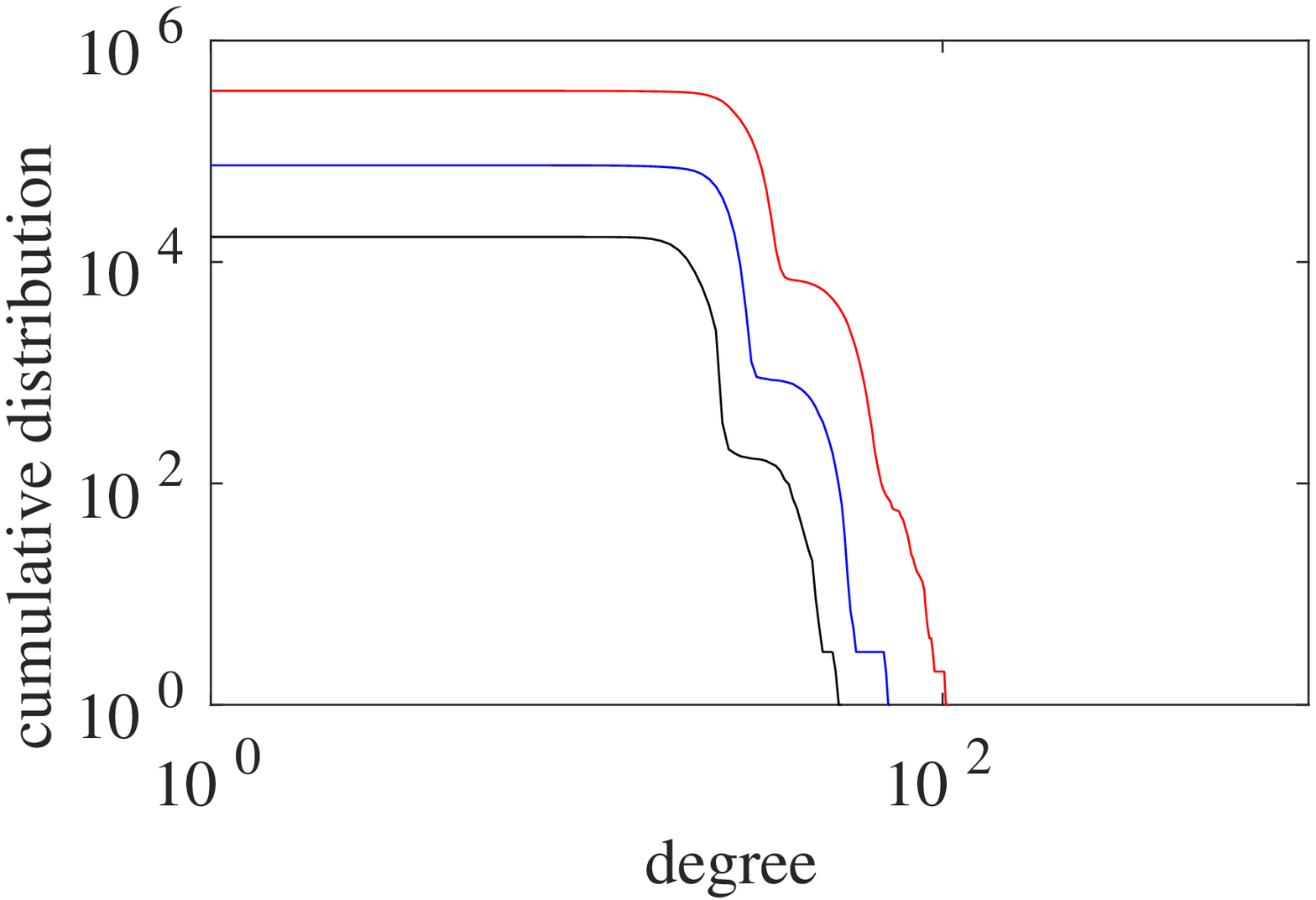}
  \label{fig:degree-dist4x4swap}
}
\quad
\subfloat[]{
  \centering
  \includegraphics[scale=.28]{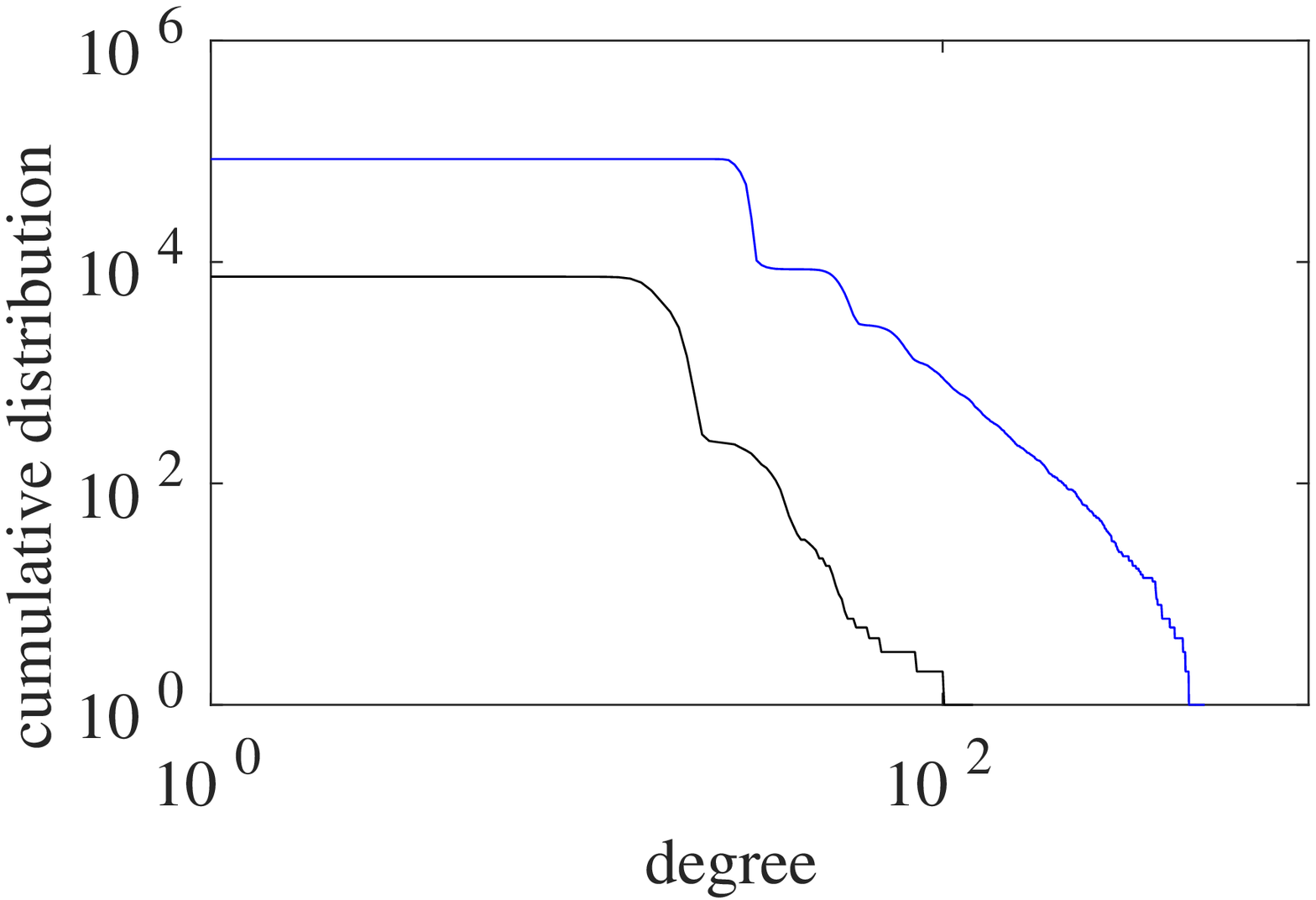}
  \label{fig:degree-dist5x5swap}
}
\quad
\subfloat[]{
  \centering
  \includegraphics[scale=.28]{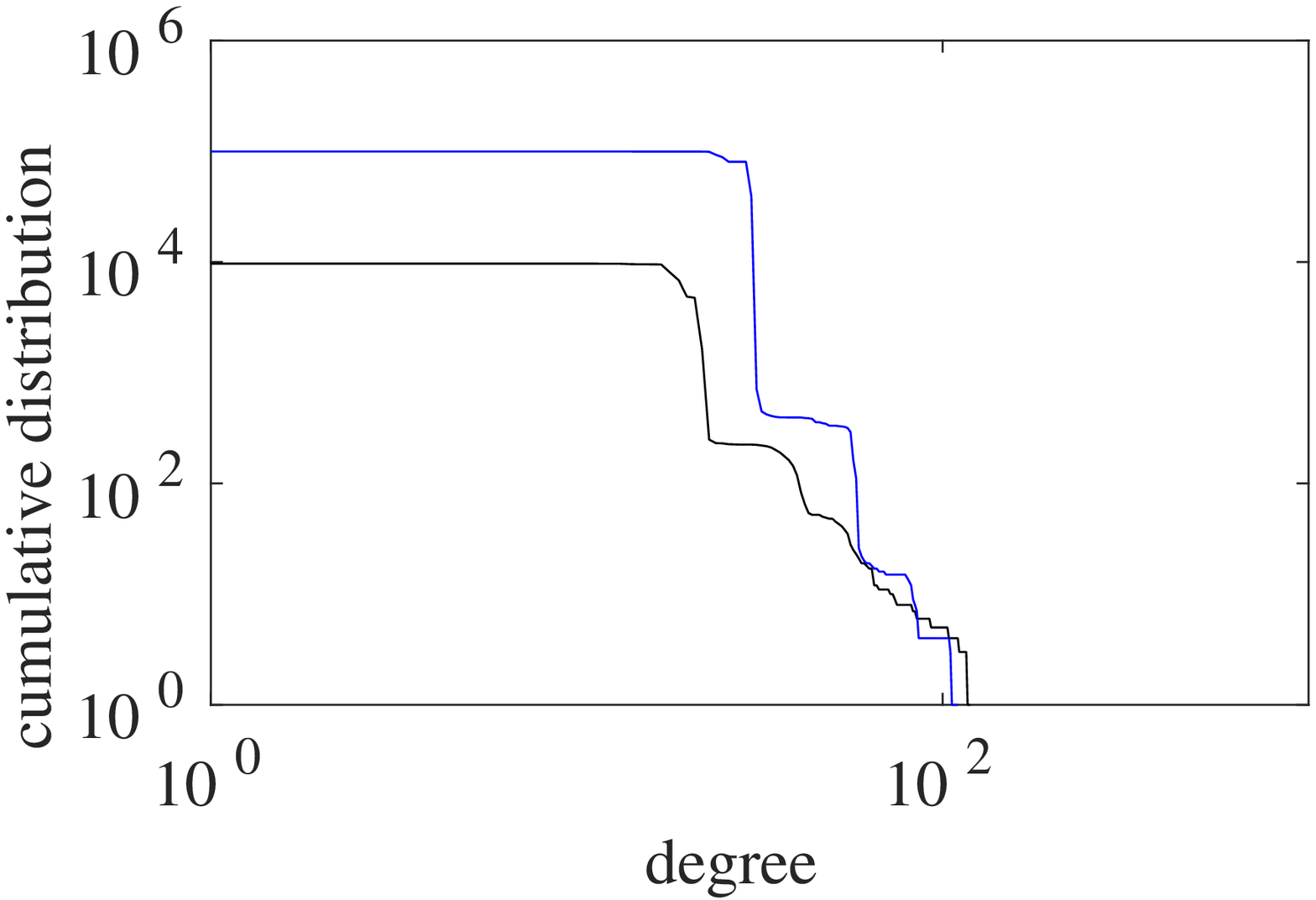}
  \label{fig:degree-dist6x6swap}
}
\caption{
Cumulative degree distribution of $NL_f$ swap for 10\,000 samples (black), 100\,000 samples (blue) and 500\,000 samples (red, when available) for a) 4x4 (20\,000 samples in black case), b) 5x5 and c) 6x6.  All curves are shown in a log-log scale.
}
\label{fig:degree-distswap}
\end{figure*}

\begin{figure}[h!]
\centering
\subfloat[]{
  \centering
  \includegraphics[scale=.35]{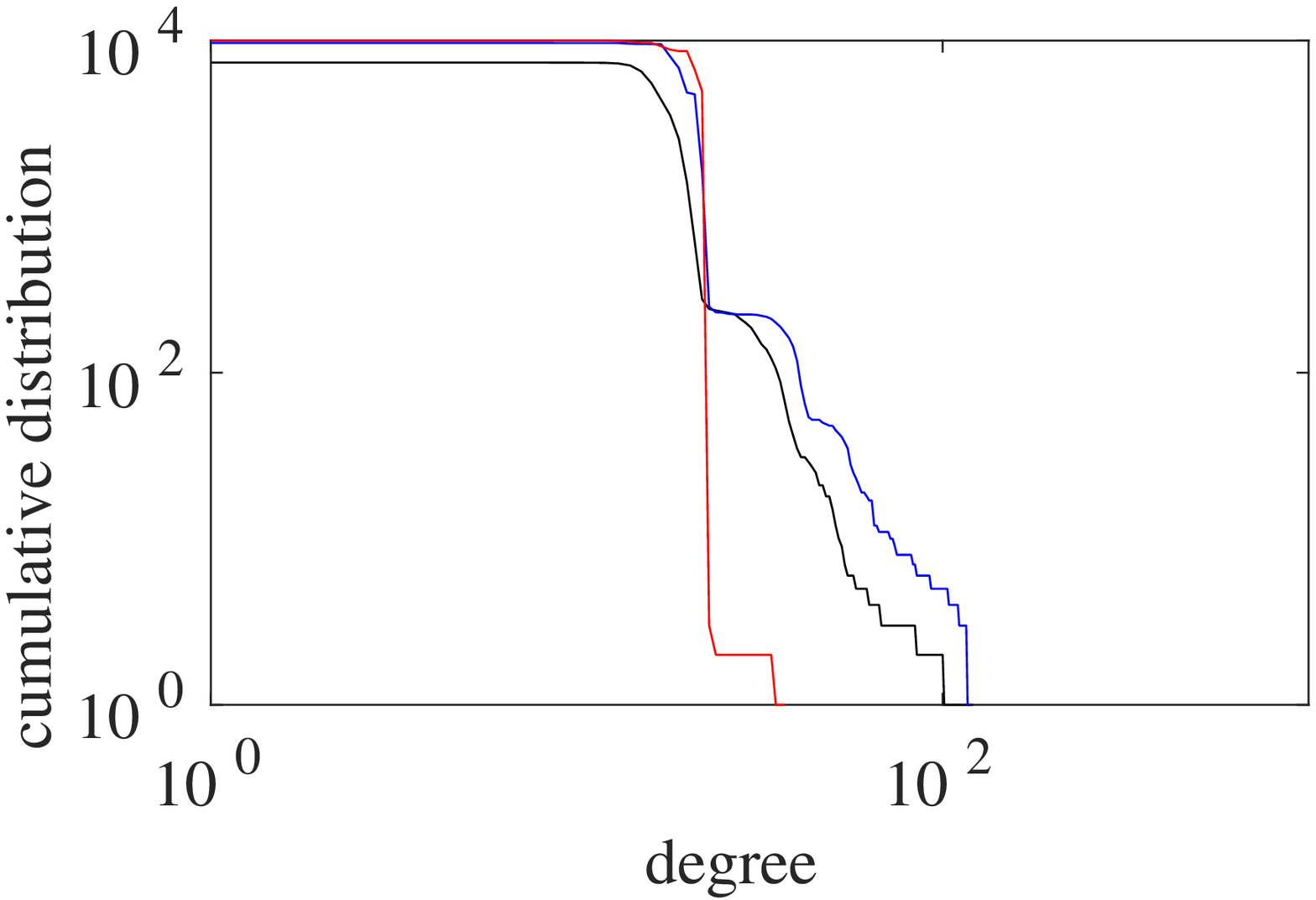}
}
\label{fig:10000swap_degree}
\quad
\subfloat[]{
  \centering
  \includegraphics[scale=.35]{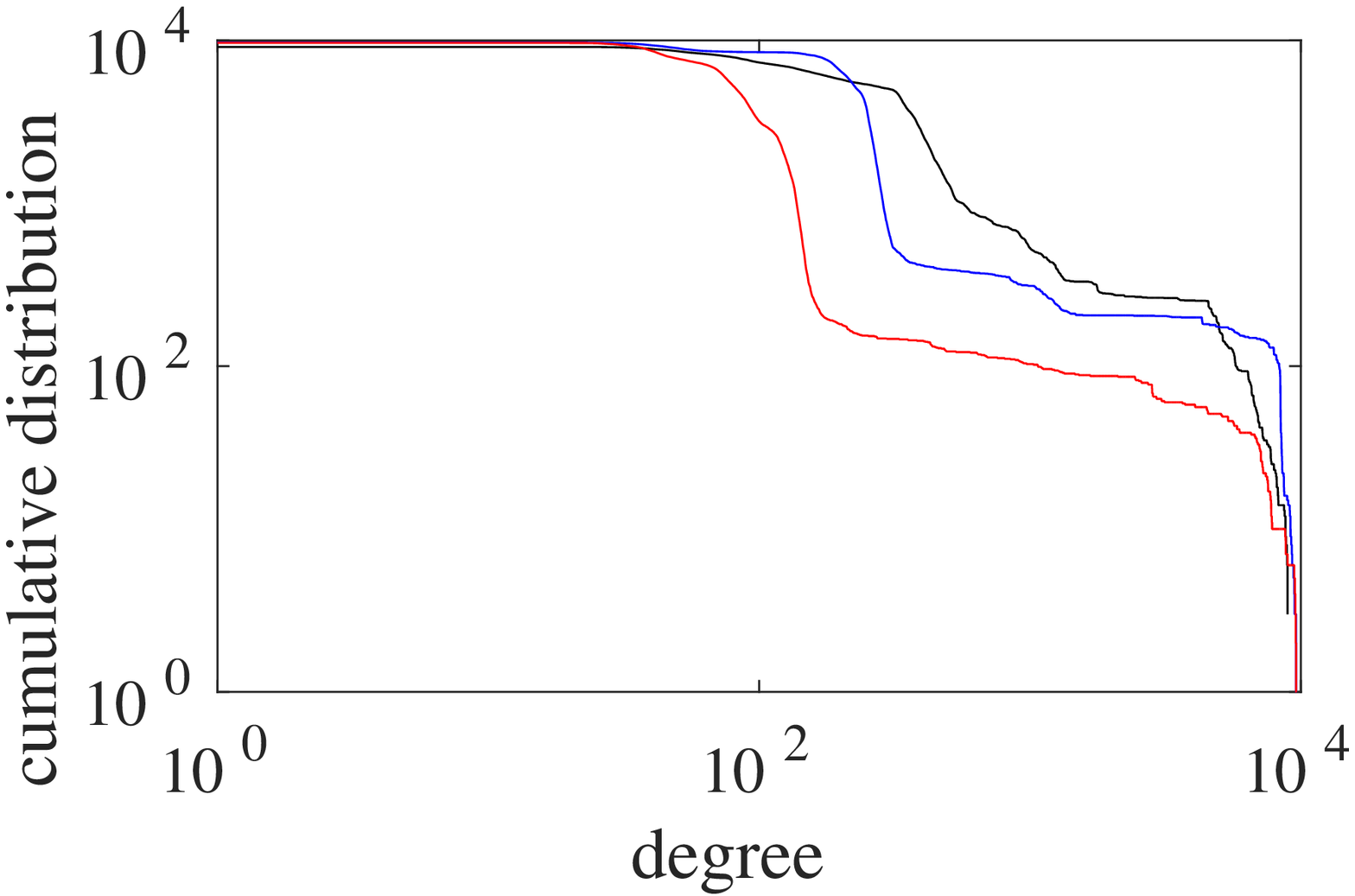}
  \label{fig:10000invert_degree}
}
\caption{Cumulative degree distribution of $NL_f$ for 5x5 (black), 6x6 (blue) and 7x7 (red) S-boxes with 10\,000 samples for a) \textit{swap} and b) \textit{invert}. All curves are shown in a log-log scale.
}\label{fig:degree:invertswap}
\end{figure}

Next, we characterise the networks visually. 
First, Figure~\ref{fig:degree-dist3x3} shows the cumulative degree distribution for 3x3 for both fitness functions and neighbourhood type, performed by an exhaustive way. Figures ~\ref{fig:degree-distswap} and 
\ref{fig:degree:invertswap} present the cumulative degree distribution for 5x5, 6x6 and 7x7, considering different number of samples.

The cumulative degree distribution function represents the probability $P(k)$ that a randomly chosen node has a degree larger than or equal to $k$.

In Figure~\ref{fig:degree-dist3x3} we can see that the degree distributions hardly decay for small degrees, specially for \textit{swap} type and for both fitness functions, while their dropping rate is significantly faster for high degrees, presenting long tails to the right for \textit{invert} type. 
This behaviour indicates that there are few nodes with large number of neighbors, however, the majority of local optima have a small number of connections.

We can note, in Figure~\ref{fig:degree-distswap}, that the dropping rate is also significantly faster for high degrees, specially for lower number of samples.

Interestingly, for 4x4 case (Figure \ref{fig:degree-dist4x4swap}) the distance from the black distributions to the blue one (factor 5 increase in samples) is roughly equivalent to the distance between the blue one and the red one (another factor 5 increase in samples) — in particular, this applies (roughly) to both dimensions in Figure 5. If the increase would be limited to a shift in the y-axis, then this would mean that the 25-fold increase in samples does not uncover different structures (as expressed in different degree distributions) in the landscape. However, the increase along the x-axis means that the rate of uncovering new structures is relatively stable -— it even increases slightly for nodes of high degree, as visible by the larger distance along the x-axis. We conjecture that the number of samples has yet to be further increased as the degree distributions does not show signs of convergence yet. At $10,000$/$20,000$ samples, the LONs for 4x4, 5x5, and 6x6 are loosely comparable, indicating some common substructures. However, at $100,000$ samples, 5x5 shows significantly more node of higher degree. This might be because $n=5$ exhibits a different substructure, or because a different substructure was not sampled in the other two cases.

Some real-world network presents these properties in the topological structure, and can be described by a power-law. or a scale-free degree distribution $P(k)=k^{-\alpha}$, where $\alpha \in[2,3]$ is a scaling parameter. 
The behaviour of local search strategies on networks has been studied according to the degree distribution~\cite{adamic200613}. The degree distribution allows one to search a power-law graph more rapidly, considering the number of edges per node varies from node to node, i.e., its edges do not let us uniformly sample the graph, but they preferentially lead to high degree nodes~\cite{Yafrani2018FLA}. 
This supports that the landscape has few nodes with high degree that efficiently connect the entire landscape, and a search at a random node has more chances to move to one of these high degree nodes, then to another node, such as an efficient way to search the entire network.

In order to study the cumulative degree distribution more rigorously, we use the Kolmogorov-Smirnov test to investigate the adequacy of power-law~\cite{clauset2009power} and exponential models~\cite{deng2011exponential}\footnote{originally proposed in \cite{ochoa2008study} to describe the degree distributions for NK models}. The test is performed on all distributions shown (see Figures~\ref{fig:degree-dist3x3},~\ref{fig:degree-distswap}, and~\ref{fig:degree:invertswap}) and with a significance level of $0.1$. The $p-values$ are presented in Table \ref{tab:KS-test}. When the $p-value>0.1$, the test fails to reject power-law and exponential as plausible distribution models. 

Table \ref{tab:KS-test} shows that the Kolmogorov-Smirnov fails to reject the the exponential plausible model for almost all the samples considered here. Only for the 3x3 LONs when using inversions, the power-law model seems to be better supported by our observations. 
Although most instances are able to produce LONs with degree distributions that fit an exponential distribution, it can not be generalised as a plausible model to describe the degree distribution for all S-box instances.

\begin{table}[h!]
\scriptsize
\centering
\begin{tabularx}{8cm}{*{7}{Y}}
\toprule
\textbf{Size}    & \textbf{Function} &\textbf{Operator} & \textbf{Samples} & \textbf{Power-Law} & \textbf{Exponential} \\ \hline

\multirow{4}{*}{3x3}\ignore{3x3NLswap}&	\multirow{2}{*}{$	NL	$}&$	swap	$&  &$	0.0321	$&$	0.3216	$\\
&\ignore{3x3NLinvert	&} &$	invert	$& &$	0.3125	$&$	0.0865	$\\
\ignore{3x3NLfswap	&}&\multirow{2}{*}{$	NL_f$}&$	swap	$&  &$	0.0364	$&$	0.4511	$\\
&\ignore{3x3Nlfinvert	&}&$	invert	$& &$	0.2658	$&$	0.0954	$\\
\midrule
\multirow{4}{*}{4x4}\ignore{4x4NLfswap20000	    }&$	Nlf	$&$	swap	$&$20,000$&$	0.0854	$&$	0.1654	$\\
\ignore{\multirow{3}{*}{4x4}4x4NLfswap100000	}&$	Nlf	$&$	swap	$& $100,000$&$	0.0954	$&$	0.1547	$\\ 
\ignore{4x4NLfswap500000	}&$	Nlf	$&$	swap	$& $500,000$ &$	0.0460	$&$	0.3215	$\\
\ignore{4x4NLfinvert100000	}&$	Nlf	$&$	invert	$& $100,000$ &$	0.0654	$&$	0.1234	$\\
\midrule
\multirow{3}{*}{5x5}\ignore{5x5NLfswap10000	    }&$	Nlf	$&$	swap	$& $10,000$ &$	0.0321	$&$	0.1325	$\\
\ignore{5x5NLfswap100000	}&$	Nlf	$&$	swap	$& $100,000$&$	0.0647	$&$	0.1795	$\\ 
\ignore{5x5NLfinvert10000	}&$	Nlf	$&$	invert	$& $10,000$ &$	0.0325	$&$	0.2154	$ \\\hline
\multirow{3}{*}{6x6}\ignore{6x6NLfswap10000	    }&$	NL_f	$&$	swap	$& $10,000$ &$	0.0990	$&$	0.2178	$\\
\ignore{6x6NLfswap100000	}&$	Nlf	$&$	swap	$& $100,000$&$	0.0217	$&$	0.3154	$\\ 
\ignore{6x6NLfinvert10000	}&$	Nlf	$&$	invert	$& $10,000$ &$	0.0645	$&$	0.3165	$\\
\midrule
\multirow{2}{*}{7x7}\ignore{7x7NLfswap10000     }&$  NL_f $&$ swap    $& $10,000$ &$	0.0548	$&$	0.2981	$\\
\ignore{7x7NLfinvert10000	}&$	Nlf	$&$	invert	$& $10,000$ &$	0.0487	$&$	0.3152	$\\
\bottomrule
\end{tabularx}
\caption{
The p-values for the Kolmogorov-Smirnov hypothesis test  with a significance level of $0.1$. If $p-value>0.1$, the test fails to reject power-law and exponential as plausible distribution models. 
}
\label{tab:KS-test}
\end{table}

\subsection{Basin of Attraction}

Exponential degree distributions do not provide a straightforward interpretation of the local search strategies' behaviours as the power-law does. Therefore, another way to analyse the difficulty of the search space for the heuristics is to consider the size of the basins of attraction.

\begin{figure*}[h]
\centering
\subfloat[]{
  \centering
  \includegraphics[scale=.20]{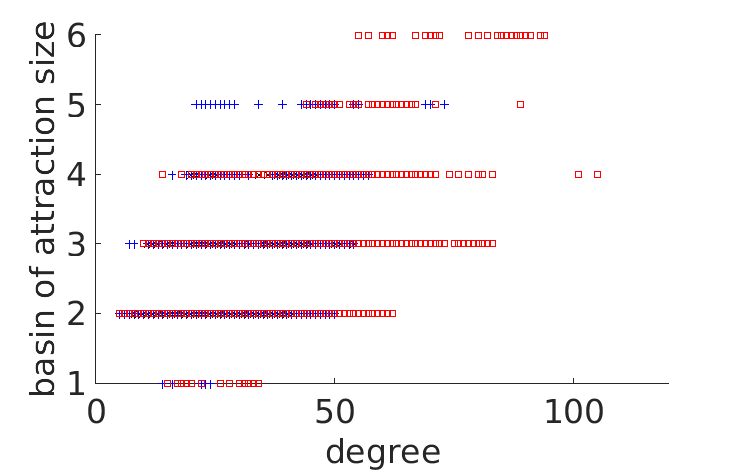}
}
\label{fig:degree-basin-4x4swap}
\,
\subfloat[]{
  \centering
  \includegraphics[scale=.20]{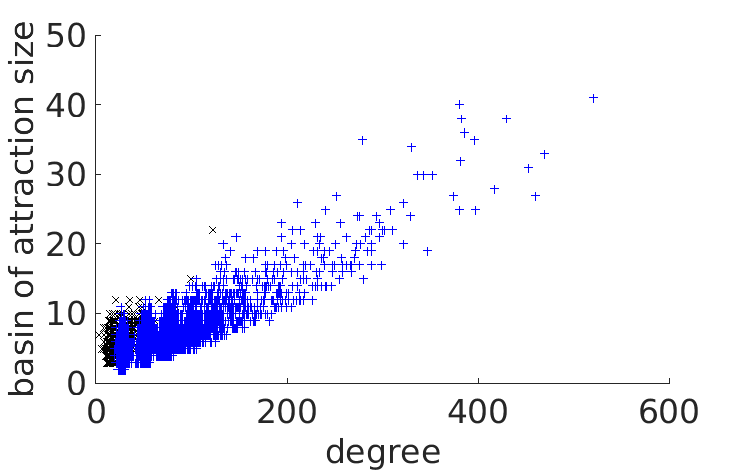}
}
\label{fig:degree-basin-5x5swap}
\,
\subfloat[]{
  \centering
  \includegraphics[scale=.20]{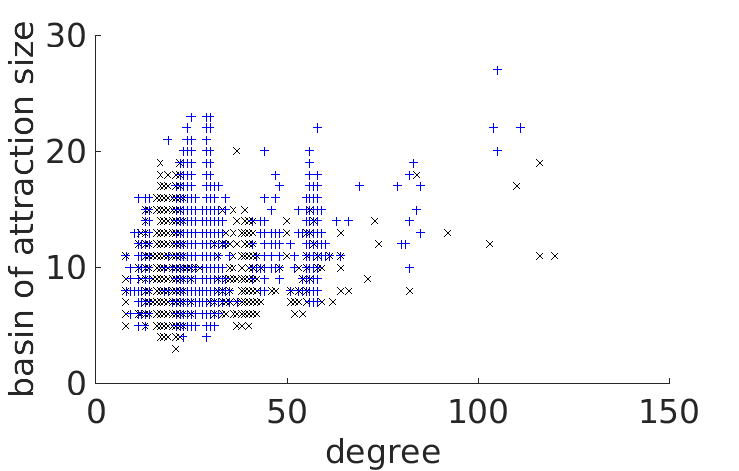}
  \label{fig:degree-basin-6x6swap}
}
\,
\subfloat[]{
  \centering
  \includegraphics[scale=.20]{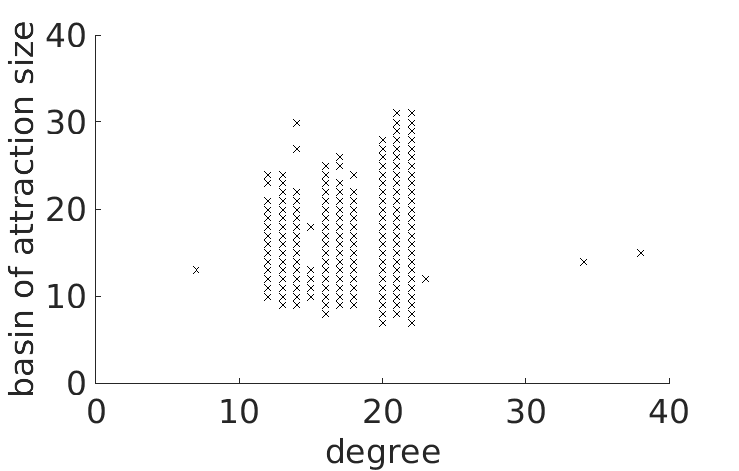}
  \label{fig:degree-basin-7x7swap}
}
\caption{
Correlation between the degree of local optima and their corresponding basin sizes on $NL_f$ swap for 10\,000 samples (black), 100\,000 samples (blue) and 500\,000 samples (red) for a) 4x4 (20\,000 samples in black case), b) 5x5 , c) 6x6 and d) 7x7. All curves are shown in a log-log scale.\label{fig:degree-basin-swap}
}
\end{figure*}

\begin{figure*}[h]
\centering
\subfloat[]{
  \centering
  \includegraphics[scale=.20]{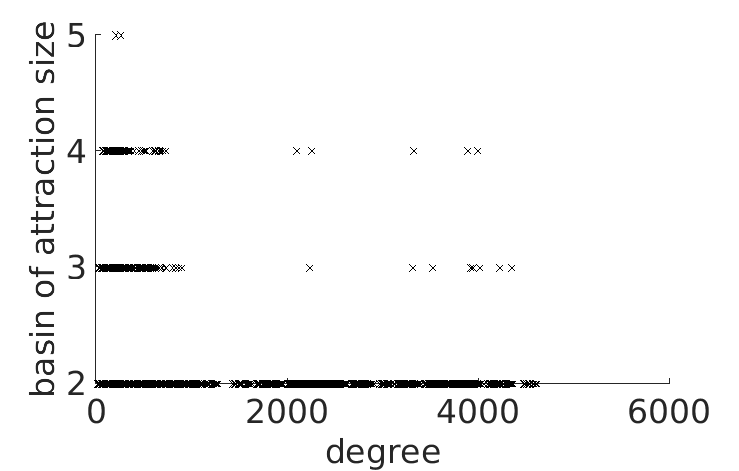}
}
\label{fig:degree-basin-4x4invert}
\,
\subfloat[]{
  \centering
  \includegraphics[scale=.20]{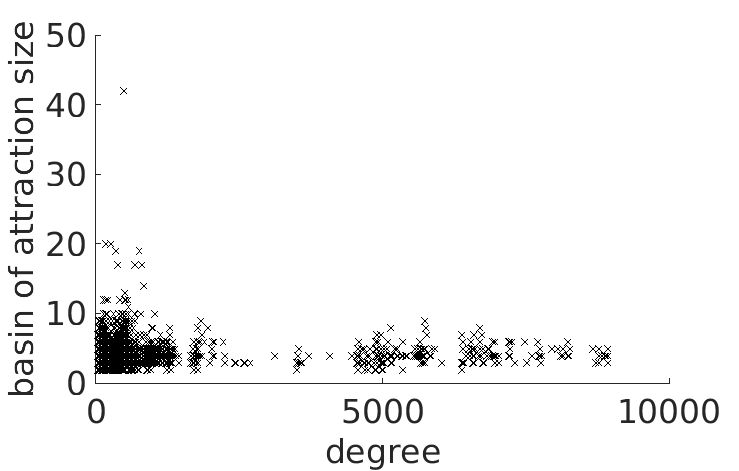}
}
\label{fig:degree-basin-5x5invert}
\,
\subfloat[]{
  \centering
  \includegraphics[scale=.20]{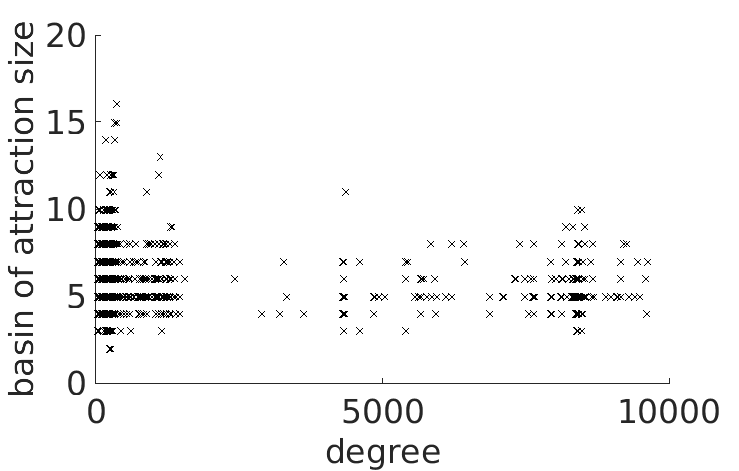}
  \label{fig:degree-basin-6x6invert}
}
\,
\subfloat[]{
  \centering
  \includegraphics[scale=.20]{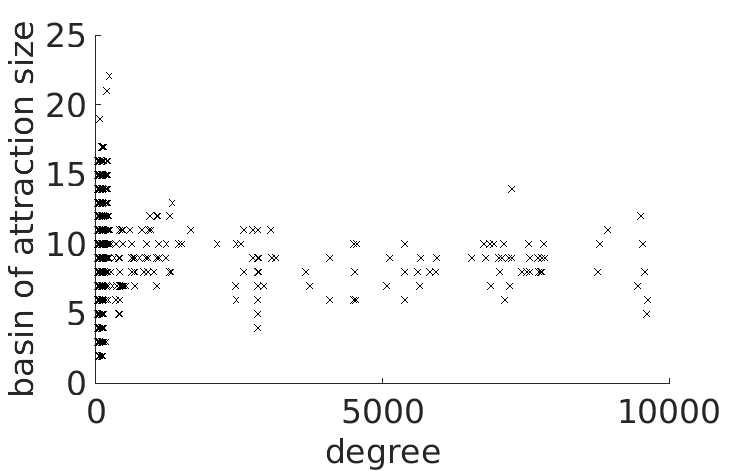}
  \label{fig:degree-basin-7x7invert}
}
\caption{
Correlation between the degree of local optima and their corresponding basin sizes on $NL_f$ invert with 10\,000 samples for a) 4x4 (100\,000 samples in this case), b) 5x5 , c) 6x6 and d) 7x7. All curves are shown in a log-log scale.\label{fig:degree-basin-invert}
}
\end{figure*}

As an example, Figures~\ref{fig:degree-basin-swap} and~\ref{fig:degree-basin-invert} show the correlation between the degree of local optima and their corresponding basin sizes when using, respectively, \textit{swap} and \textit{invert}. 
Note that we are not seeing a strict subset-superset relationship as the starting point for the lexicographic enumeration of the samples was randomly drawn.

Remarkably, in Figure~\ref{fig:degree-basin-swap}, the 5x5 LON  at $100,000$ samples is again quite different in nature from the other three, as only it contains nodes of degrees up to 500, whereas the others contains nodes of degrees up to 120 only. All three LONs at $10,000$ samples can be considered somewhat comparable, with many nodes of degree 20--50 and small basins of attraction of size 2--20. However, regarding Figure~\ref{fig:degree-basin-invert}, we note nodes with high degree (up to $10,000$ due to neighbourhood-relationships with other local optima) and also small basins of attraction size. This indicates the existence of large plateaus, where each point on the plateau has its own, small basin of attraction.

If this difference is true and not just based on a lucky sample, then this shows that the S-boxes with odd $n$ are significantly different problems (at least when using \textit{swap}, as we have seen high-degree nodes (with degrees up to 200) in the 3x3 case as well. Note, however, that this does not carry over to the \textit{invert} case, see Figure~\ref{fig:degree-basin-invert}.

Independent of this, we can see a correlation between the degrees and the basin sizes, suggesting that instances can be searched more effectively due to they present many neighbours belonging to the same large basin of attraction.

\begin{figure*}[h]
\centering
\subfloat[]{
  \centering
  \includegraphics[scale=.19]{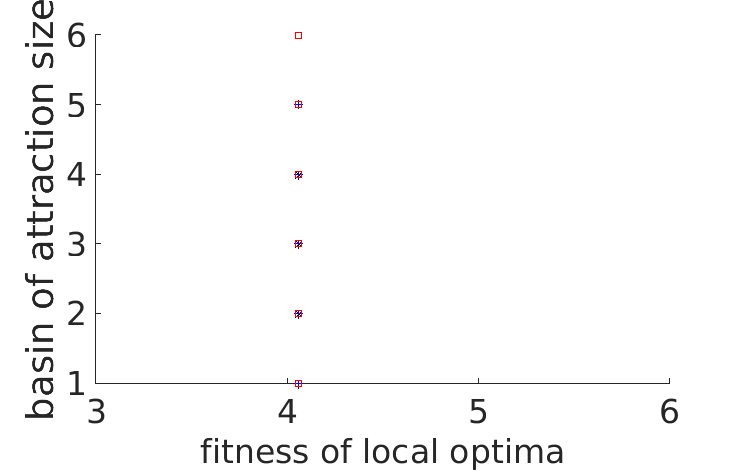}
}
\label{fig:basin-lofitness4x4}
\quad
\subfloat[]{
  \centering
  \includegraphics[scale=.19]{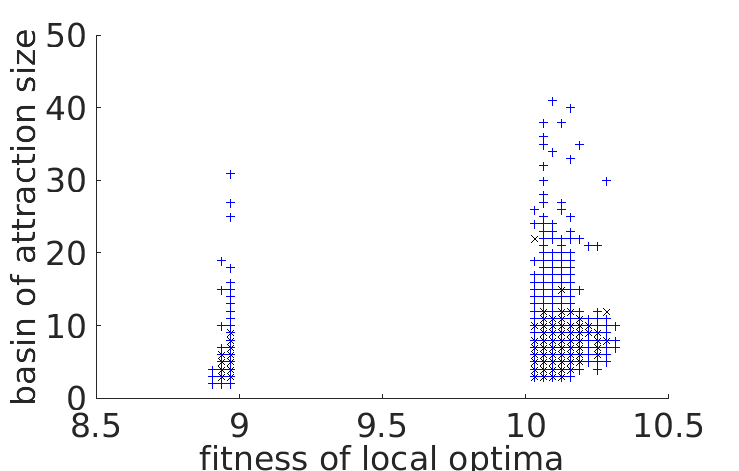}
}
\label{fig:basin-lofitness5x5}
\quad
\subfloat[]{
  \centering
  \includegraphics[scale=.19]{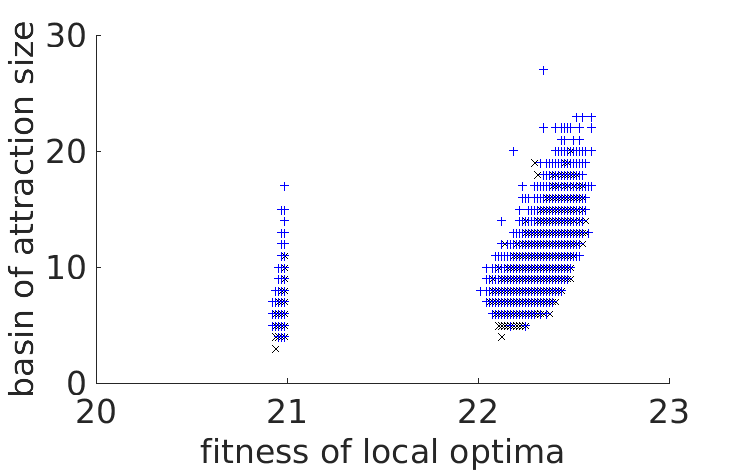}
  \label{fig:basin-lofitness6x6}
}
\quad
\subfloat[]{
  \centering
  \includegraphics[scale=.19]{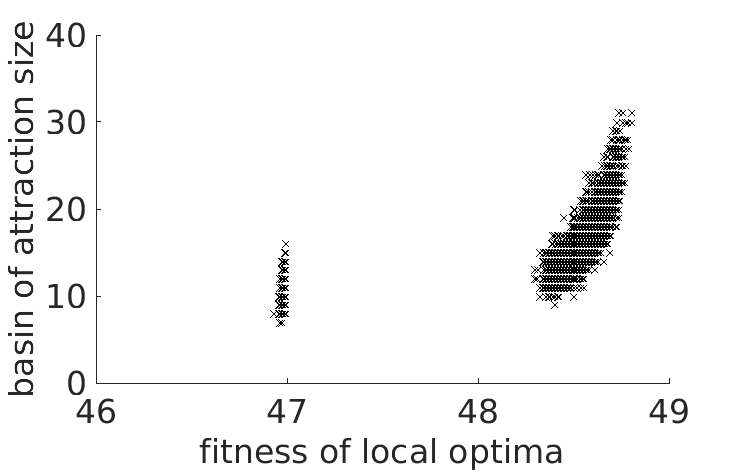}
  \label{fig:basin-lofitness7x7}
}
\caption{
Correlation between the fitness of local optima and their corresponding basin sizes on $NL_f$ swap for 10\,000 samples (black), 100\,000 samples (blue) and 500\,000 samples 
(red) for a) 4x4 (20\,000 samples in black case), b) 5x5 
c) 6x6 and d) 7x7. Note that none of the almost one million greedy searches has returned an optimal solution.\label{fig:basin-lofitness}
}
\end{figure*}

\begin{figure*}[h]
\centering
\subfloat[]{
  \centering
  \includegraphics[scale=.19]{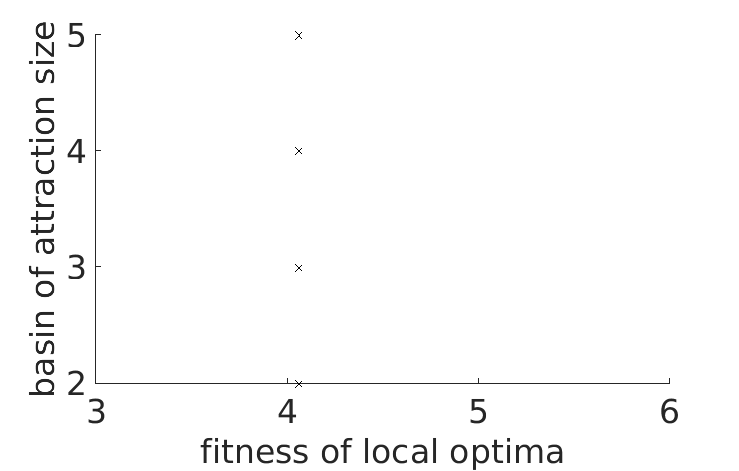}
}
\label{fig:basin-lofitnessinvert4x4}
\quad
\subfloat[]{
  \centering
  \includegraphics[scale=.19]{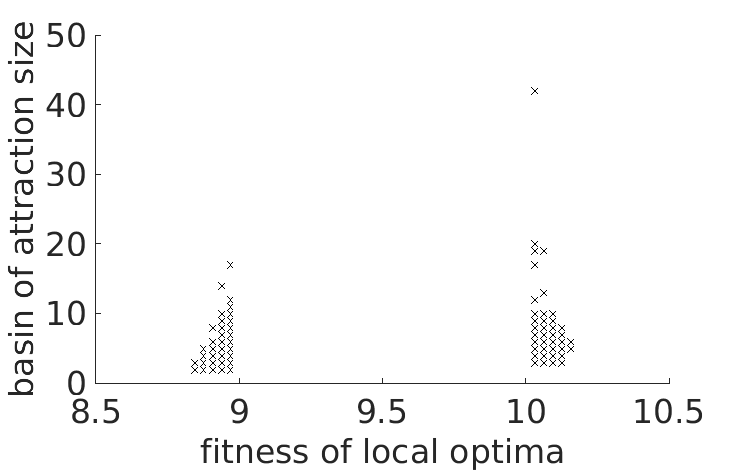}
}
\label{fig:basin-lofitnessinvert5x5}
\quad
\subfloat[]{
  \centering
  \includegraphics[scale=.19]{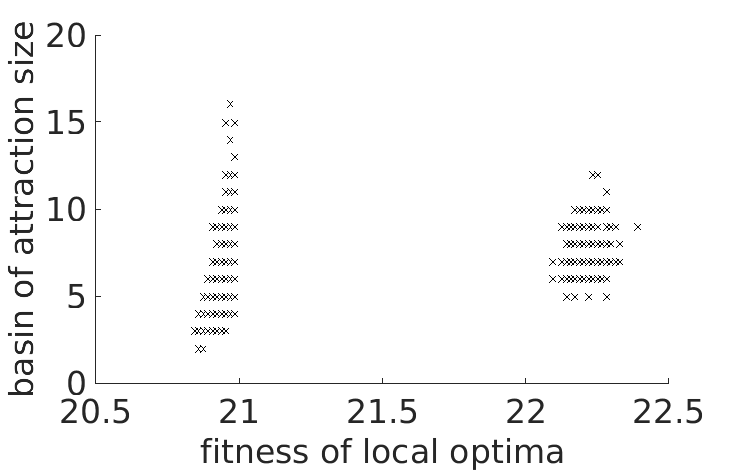}
  \label{fig:basin-lofitnessinvert6x6}
}
\quad
\subfloat[]{
  \centering
  \includegraphics[scale=.19]{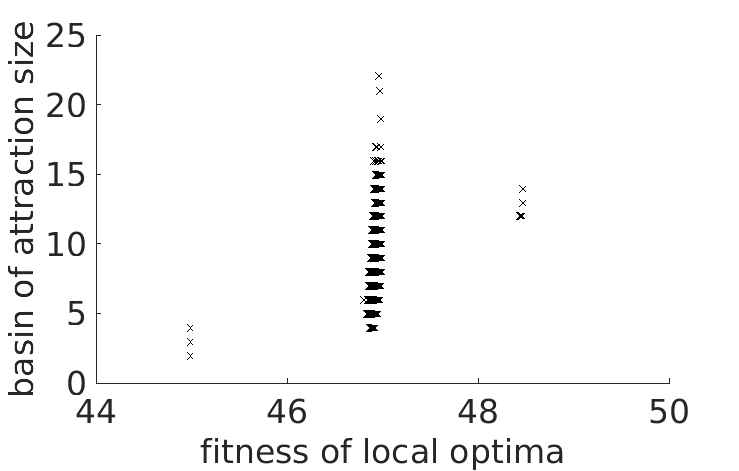}
  \label{fig:basin-lofitnessinvert7x7}
}
\caption{
Correlation between the fitness of local optima and their corresponding basin sizes on $NL_f$ invert with 10\,000 samples (black), for a) 4x4 (100\,000 samples in this case), b) 5x5 
c) 6x6 and d) 7x7. \label{fig:basin-lofitness-invert}
}
\end{figure*}

Lastly, we show in Figures~\ref{fig:basin-lofitness} and \ref{fig:basin-lofitness-invert} the relationship of the fitness of a local optimum and the corresponding basin size for \textit{swap} and \textit{invert}, respectively. As there are many small basins, and also many basins of comparable fitness, we conjecture that navigating this landscape is hard, as there is little information that a search heuristic could latch onto. For the 6x6 case (Figure \ref{fig:basin-lofitness6x6}), we can observe a correlation of basin size and fitness. It should be possible to estimate this characteristic in a search heuristic with restarts: if a local search returns a solution of poor fitness, then a not-too-small perturbation should be done to determine the starting point for the next run, in order to increase chances to escape the small, bad basin of attraction. Note that the opposite does not hold, and a large perturbation is not a guarantee for success.

\section{Conclusions}
\label{sec:conclusions}

With this article, we present the first fitness-landscape analysis of S-boxes for cryptographic uses -- to the best of our knowledge, this is even the first landscape analysis in the greater field of security. 

We find that almost every single one of the about 1 million conducted runs by our local search finds a different local optimum. This is the consequence of many small basins of attraction, even though the local optima networks themselves are very dense. These and additional insights gained will inform our algorithm design in the future. On the experimental side, future studies will include combinations of Tabu lists or niching approaches with restarts where the impact of the perturbation from the previous starting point can be controlled.

Our Interesting results for the degree of nodes when considering odd and even cases warrant further investigation, ideally coupled with theoretical results. S-boxes of odd sizes are able to reach the Sidelnikov-Chabaud-Vaudenay bound, which means that the maximal values that can appear in the Walsh-Hadamard spectrum have the same magnitude as one size smaller S-box that is in even dimension. This means there is less variability for odd dimensions, which would indicate a more difficult problem for heuristics.

\begin{acks}
Our work was supported by the Australian Research Council project DE160100850.
\end{acks}
\balance


\end{document}